\documentclass{article} 
\usepackage{iclr2017_conference,times}
\usepackage{hyperref}
\usepackage{url}
\usepackage{graphicx}
\usepackage{tabulary}
\usepackage{makecell}
\usepackage{subfigure}
\newcommand{\etal}{\textit{et al}.$~$}
\renewcommand{\cite}{\citep}
\def\vsp{\vspace{-0.15in}}
\newcommand{\cb}{\centering\arraybackslash}

\title{SqueezeNet: AlexNet-level accuracy with 50x fewer parameters and $<$0.5MB model size}

\author{
	{\makecell[l]{Forrest N. Iandola$^{1}$, Song Han$^2$, Matthew W. Moskewicz$^{1}$, Khalid Ashraf$^{1}$, \\ William J. Dally$^2$, Kurt Keutzer$^1$}} \\
	{ $^1$DeepScale\thanks{\href{http://deepscale.ai}{http://deepscale.ai}} ~ \& UC Berkeley ~~~~ $^2$Stanford University} \\
	\makecell[l]{\tt\footnotesize \{forresti, moskewcz, kashraf, keutzer\}@eecs.berkeley.edu \\ \tt\footnotesize \{songhan, dally\}@stanford.edu}
}

\begin{document}
	
\maketitle

\begin{abstract}
Recent research on deep convolutional neural networks (CNNs) has focused primarily on improving accuracy.
For a given accuracy level, it is typically possible to identify multiple CNN architectures that achieve that accuracy level.
With equivalent accuracy, smaller CNN architectures offer at least three advantages:
(1) Smaller CNNs require less communication across servers during distributed training.
(2) Smaller CNNs require less bandwidth to export a new model from the cloud to an autonomous car.
(3) Smaller CNNs are more feasible to deploy on FPGAs and other hardware with limited memory.
To provide all of these advantages, we propose a small CNN architecture called SqueezeNet.
SqueezeNet achieves AlexNet-level accuracy on ImageNet with 50x fewer parameters.
Additionally, with model compression techniques, we are able to compress SqueezeNet to less than 0.5MB ($510\times$ smaller than AlexNet).

The SqueezeNet architecture is available for download here:
\href{https://github.com/DeepScale/SqueezeNet}{https://github.com/DeepScale/SqueezeNet}
\end{abstract}
	

%

\section{Introduction and Motivation}
\vsp
\label{sec:intro-ch5}
Much of the recent research on deep convolutional neural networks (CNNs) has focused on increasing accuracy on computer vision datasets.
For a given accuracy level, there typically exist multiple CNN architectures that achieve that accuracy level.
Given equivalent accuracy, a CNN architecture with fewer parameters has several advantages:
\begin{itemize}
	\setlength\itemsep{0in} 
	\item[$\bullet$]{{\bf More efficient distributed training.} 
		Communication among servers is the limiting factor to the scalability of distributed CNN training.
		For distributed data-parallel training, communication overhead is directly proportional to the number of parameters in the model~\cite{FireCaffe}.
		In short, small models train faster due to requiring less communication.
	}
	
	\item[$\bullet$]{{\bf Less overhead when exporting new models to clients.} For autonomous driving, companies such as Tesla periodically copy new models from their servers to customers' cars. This practice is often referred to as an {\em over-the-air} update. Consumer Reports has found that the safety of Tesla's {\em Autopilot} semi-autonomous driving functionality has incrementally improved with recent over-the-air updates~\cite{ConsumerReports-Tesla}. However, over-the-air updates of today's typical CNN/DNN models can require large data transfers. With AlexNet, this would require 240MB of communication from the server to the car. Smaller models require less communication, making frequent updates more feasible.}
	
	\item[$\bullet$]{{\bf Feasible FPGA and embedded deployment.} FPGAs often have less than 10MB\footnote{For example, the Xilinx Vertex-7 FPGA has a maximum of 8.5 MBytes (i.e. 68 Mbits) of on-chip memory and does not provide off-chip memory.} of on-chip memory and no off-chip memory or storage. 
		For inference, a sufficiently small model could be stored directly on the FPGA instead of being bottlenecked by memory bandwidth~\cite{fpga2016cnn}, while video frames stream through the FPGA in real time.
		Further, when deploying CNNs on Application-Specific Integrated Circuits (ASICs), a sufficiently small model could be stored directly on-chip, and smaller models may enable the ASIC to fit on a smaller die.}
	%
	
	
\end{itemize}

As you can see, there are several advantages of smaller CNN architectures.
With this in mind, we focus directly on the problem of identifying a CNN architecture with fewer parameters but equivalent accuracy compared to a well-known model.
We have discovered such an architecture, which we call {\em SqueezeNet.}
In addition, we present our attempt at a more disciplined approach to searching the design space for novel CNN architectures.

The rest of the paper is organized as follows. 
In Section~\ref{sec:related} we review the related work.
Then, in Sections~\ref{sec:SqueezeNet} and~\ref{sec:eval-squeezenet} we describe and evaluate the SqueezeNet architecture. 
After that, we turn our attention to understanding how CNN architectural design choices impact model size and accuracy.
We gain this understanding by exploring the design space of SqueezeNet-like architectures.
In Section~\ref{sec:microDSE}, we do design space exploration on the {\em CNN microarchitecture}, which we define as the organization and dimensionality of individual layers and modules.
In Section~\ref{sec:macroDSE}, we do design space exploration on the {\em CNN macroarchitecture}, which we define as high-level organization of layers in a CNN.
Finally, we conclude in Section~\ref{sec:conclusions}.
In short, Sections~\ref{sec:SqueezeNet} and~\ref{sec:eval-squeezenet} are useful for CNN researchers as well as practitioners who simply want to apply SqueezeNet to a new application.
The remaining sections are aimed at advanced researchers who intend to design their own CNN architectures. 

\section{Related Work}
\label{sec:related}
\vsp

\subsection{Model Compression}
\label{sec:related-model-compression}
\vsp

The overarching goal of our work is to identify a model that has very few parameters while preserving accuracy.
To address this problem, a sensible approach is to take an existing CNN model and compress it in a lossy fashion.
In fact, a research community has emerged around the topic of {\em model compression}, and several approaches have been reported.
A fairly straightforward approach by Denton \etal is to apply singular value decomposition (SVD) to a pretrained CNN model~\cite{facebook-compress-2014}.
Han \etal developed Network Pruning, which begins with a pretrained model, then replaces parameters that are below a certain threshold with zeros to form a sparse matrix, and finally performs a few iterations of training on the sparse CNN~\cite{dally2015-1}.
Recently, Han \etal extended their work by combining Network Pruning with quantization (to 8 bits or less) and huffman encoding to create an approach called Deep Compression~\cite{dally2015-2}, and further designed a hardware accelerator called EIE~\cite{EIE} that operates directly on the compressed model, achieving substantial speedups and energy savings.

\subsection{CNN Microarchitecture}
\vsp

Convolutions have been used in artificial neural networks for at least 25 years; LeCun \etal helped to popularize CNNs for digit recognition applications in the late 1980s~\cite{LeCun89}.
In neural networks, convolution filters are typically 3D, with height, width, and channels as the key dimensions.
When applied to images, CNN filters typically have 3 channels in their first layer (i.e. RGB), and in each subsequent layer $L_i$ the filters have the same number of channels as $L_{i-1}$ has filters.
The early work by LeCun \etal\cite{LeCun89} uses 5x5xChannels\footnote{From now on, we will simply abbreviate HxWxChannels to HxW.} filters, and the recent VGG~\cite{VGG-19} architectures extensively use 3x3 filters.
Models such as Network-in-Network~\cite{NiN} and the GoogLeNet family of architectures~\cite{googlenet,googleBN,googlenet-v3,googlenet-v4} use 1x1 filters in some layers.

With the trend of designing very deep CNNs, it becomes cumbersome to manually select filter dimensions for each layer.
To address this, various higher level building blocks, or {\em modules}, comprised of multiple convolution layers with a specific fixed organization have been proposed.
For example, the GoogLeNet papers propose {\em Inception modules}, which are comprised of a number of different dimensionalities of filters, usually including 1x1 and 3x3, plus sometimes 5x5~\cite{googlenet} and sometimes 1x3 and 3x1~\cite{googlenet-v3}.
Many such modules are then combined, perhaps with additional {\em ad-hoc} layers, to form a complete network. 
We use the term {\em CNN microarchitecture} to refer to the particular organization and dimensions of the individual modules. 

\subsection{CNN Macroarchitecture}
\vsp

While the CNN microarchitecture refers to individual layers and modules, we define the {\em CNN macroarchitecture} as the system-level organization of multiple modules into an end-to-end CNN architecture.

Perhaps the mostly widely studied CNN macroarchitecture topic in the recent literature is the impact of {\em depth} (i.e. number of layers) in networks.
Simoyan and Zisserman proposed the VGG~\cite{VGG-19} family of CNNs with 12 to 19 layers and reported that deeper networks produce higher accuracy on the ImageNet-1k dataset~\cite{imagenet}.
K. He \etal proposed deeper CNNs with up to 30 layers that deliver even higher ImageNet accuracy~\cite{He2015}.

The choice of connections across multiple layers or modules is an emerging area of CNN macroarchitectural research.
Residual Networks (ResNet)~\cite{resnet} and Highway Networks~\cite{highway-networks} each propose the use of connections that skip over multiple layers, for example additively connecting the activations from layer 3 to the activations from layer 6.
We refer to these connections as {\em bypass connections}.
The authors of ResNet provide an A/B comparison of a 34-layer CNN with and without bypass connections; adding bypass connections delivers a 2 percentage-point improvement on Top-5 ImageNet accuracy.

\subsection{Neural Network Design Space Exploration}
\vsp

Neural networks (including deep and convolutional NNs) have a large design space, with numerous options for microarchitectures, macroarchitectures, solvers, and other hyperparameters.
It seems natural that the community would want to gain intuition about how these factors impact a NN's accuracy (i.e. the {\em shape} of the design space).
Much of the work on design space exploration (DSE) of NNs has focused on developing automated approaches for finding NN architectures that deliver higher accuracy.
These automated DSE approaches include bayesian optimization~\cite{spearmint-paper}, simulated annealing~\cite{simulated-annealing}, randomized search~\cite{randomized-search}, and genetic algorithms~\cite{genetic-algorithms}.
To their credit, each of these papers provides a case in which the proposed DSE approach produces a NN architecture that achieves higher accuracy compared to a representative baseline.
However, these papers make no attempt to provide intuition about the shape of the NN design space.
Later in this paper, we eschew automated approaches -- instead, we refactor CNNs in such a way that we can do principled A/B comparisons to investigate how CNN architectural decisions influence model size and accuracy.

In the following sections, we first propose and evaluate the SqueezeNet architecture with and without model compression. 
Then, we explore the impact of design choices in microarchitecture and macroarchitecture for SqueezeNet-like CNN architectures.

\section{SqueezeNet: preserving accuracy with few parameters}
\label{sec:SqueezeNet}
\vsp

In this section, we begin by outlining our design strategies for CNN architectures with few parameters.
Then, we introduce the {\em Fire module}, our new building block out of which to build CNN architectures.
Finally, we use our design strategies to construct {\em SqueezeNet}, which is comprised mainly of Fire modules.

\subsection{Architectural Design Strategies}
\label{sec:design-strategies}
\vsp

Our overarching objective in this paper is to identify CNN architectures that have few parameters while maintaining competitive accuracy. 
To achieve this, we employ three main strategies when designing CNN architectures: 

\noindent
{\bf {\em Strategy 1.} Replace 3x3 filters with 1x1 filters.} 
Given a budget of a certain number of convolution filters, we will choose to make the majority of these filters 1x1, since a 1x1 filter has 9X fewer parameters than a 3x3 filter. 
\vspace{0.1in} 

\noindent
{\bf {\em Strategy 2.} Decrease the number of input channels to 3x3 filters.} 
Consider a convolution layer that is comprised entirely of 3x3 filters. 
The total quantity of parameters in this layer is (number of input channels) * (number of filters) * (3*3). 
So, to maintain a small total number of parameters in a CNN, it is important not only to decrease the number of 3x3 filters (see Strategy 1 above), but also to decrease the number of {\em input channels} to the 3x3 filters. 
We decrease the number of input channels to 3x3 filters using {\em squeeze layers}, which we describe in the next section. 
\vspace{0.05in} 

\noindent
{\bf {\em Strategy 3.} Downsample late in the network so that convolution layers have large activation maps.}
In a convolutional network, each convolution layer produces an output activation map with a spatial resolution that is at least 1x1 and often much larger than 1x1. 
The height and width of these activation maps are controlled by: (1) the size of the input data (e.g. 256x256 images) and (2) the choice of layers in which to downsample in the CNN architecture.
Most commonly, downsampling is engineered into CNN architectures by setting the (stride $>$ 1) in some of the convolution or pooling layers (e.g.~\cite{googlenet,VGG-19,alexnet}).
If early\footnote{In our terminology, an ``early" layer is close to the input data.} layers in the network have large strides, then most layers will have small activation maps.
Conversely, if most layers in the network have a stride of 1, and the strides greater than 1 are concentrated toward the end\footnote{In our terminology, the ``end" of the network is the classifier.} of the network, then many layers in the network will have large activation maps.
Our intuition is that large activation maps (due to delayed downsampling) can lead to higher classification accuracy, with all else held equal.
Indeed, K. He and H. Sun applied delayed downsampling to four different CNN architectures, and in each case delayed downsampling led to higher classification accuracy~\cite{ConstrainedTimeCost}. 

\begin{figure}[!t]
	\centering
	\fbox{	
		\includegraphics[width=3.5in]{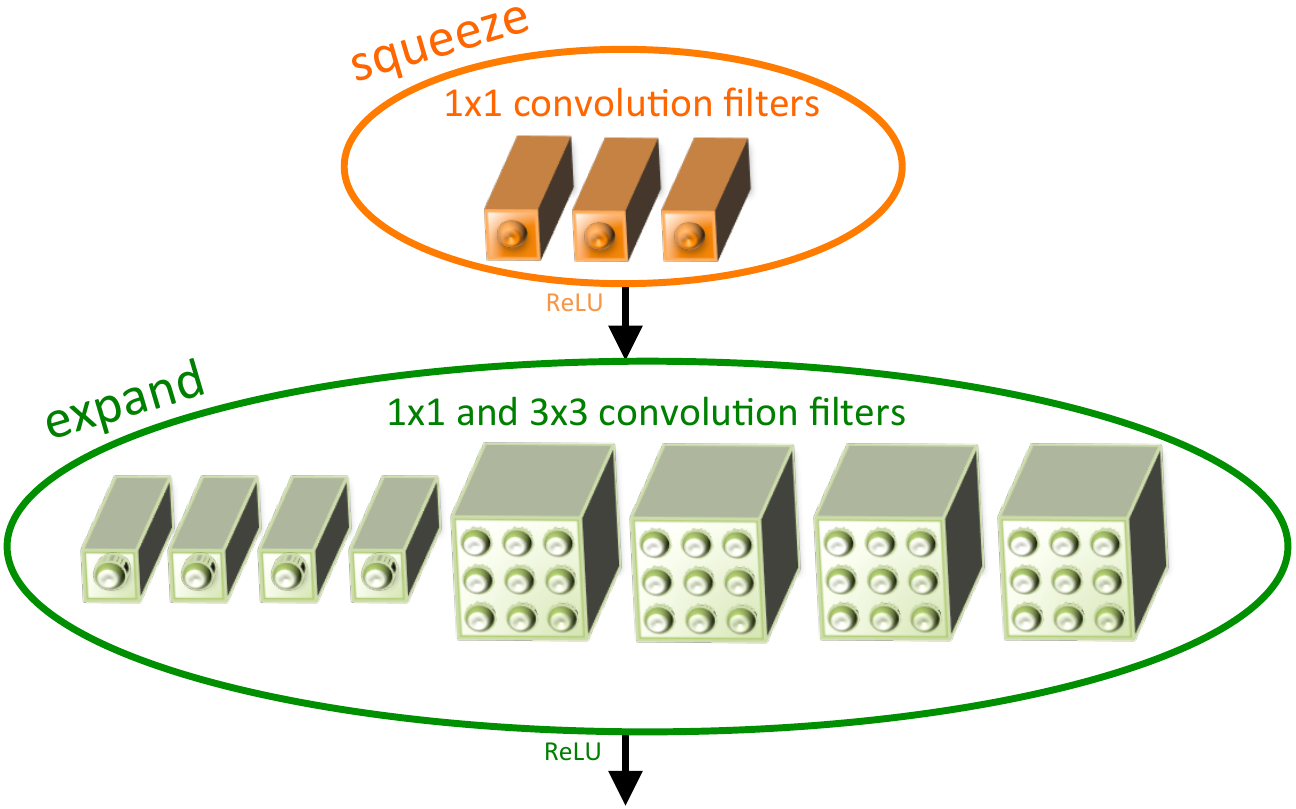}
	}
	\caption[Microarchitectural view of the Fire module]{Microarchitectural view: Organization of convolution filters in the {\bf Fire module}. In this example, $s_{1x1}=3$, $e_{1x1}=4$, and $e_{3x3}=4$. We illustrate the convolution filters but not the activations.}
	\label{fig:fire-module}
\end{figure}

Strategies 1 and 2 are about judiciously decreasing the quantity of parameters in a CNN while attempting to preserve accuracy.
Strategy 3 is about maximizing accuracy on a limited budget of parameters.
Next, we describe the Fire module, which is our building block for CNN architectures that enables us to successfully employ Strategies 1, 2, and 3.

\begin{figure*}[!t]
	\centering
	\includegraphics[height=0.616\textwidth]{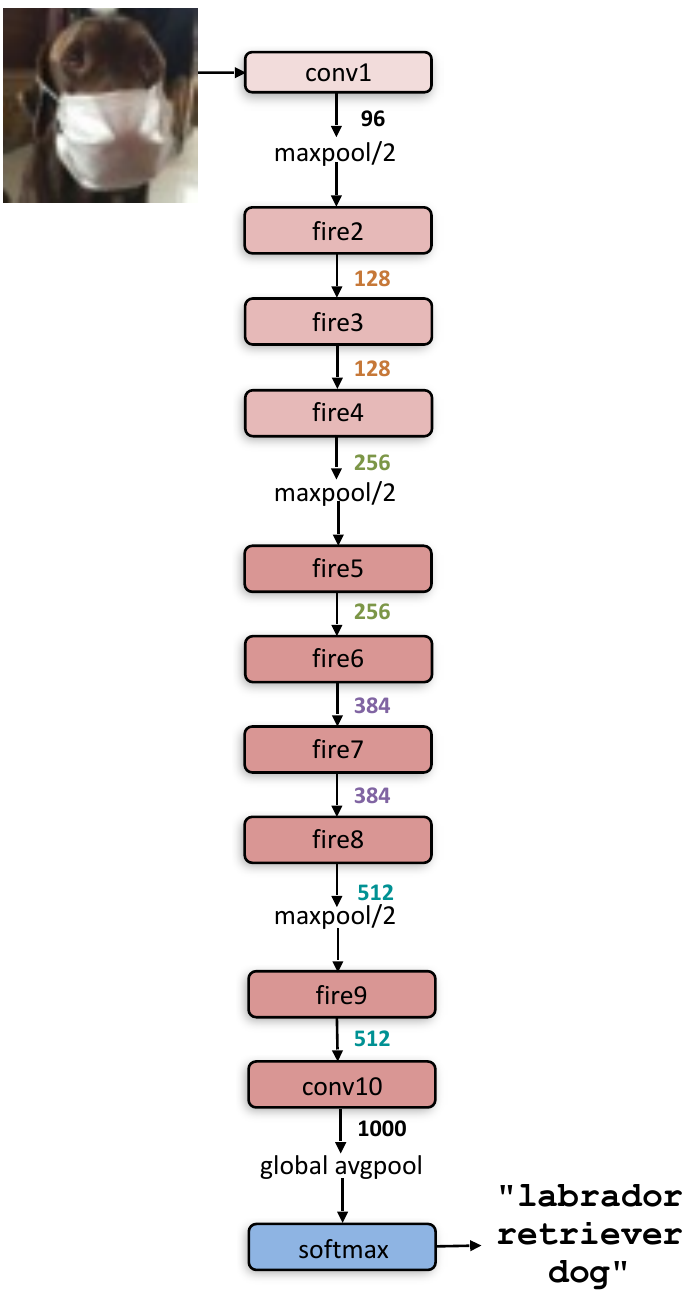}
	\hspace{1em}
	\includegraphics[height=0.6\textwidth]{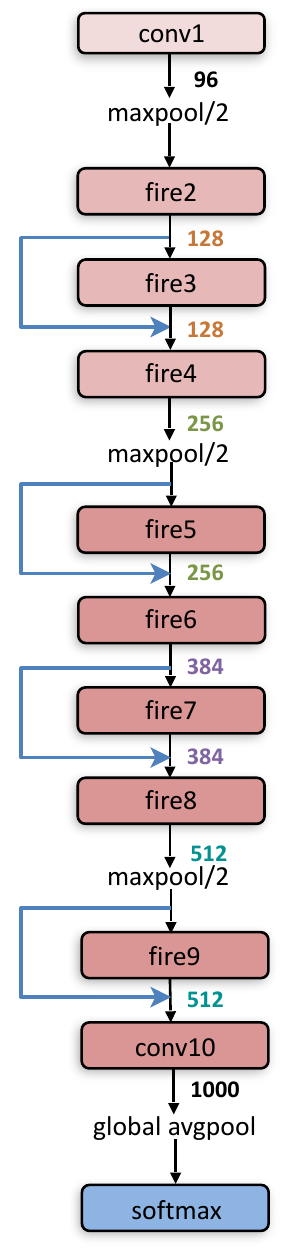}
	\hspace{5em}
	\includegraphics[height=0.6\textwidth]{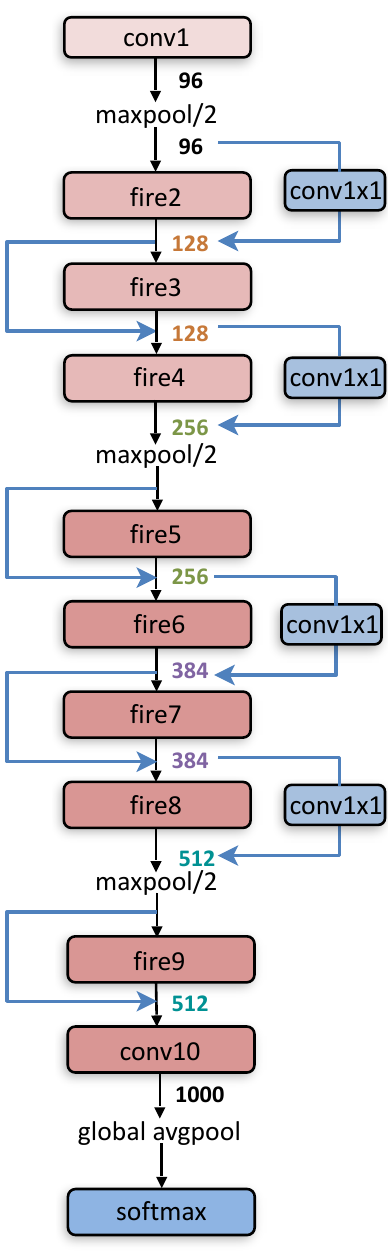}
	\caption[Macroarchitectural view of the SqueezeNet architecture]{Macroarchitectural view of our SqueezeNet architecture. Left: SqueezeNet (Section~\ref{sec:squeezenet-arch}); Middle: SqueezeNet with simple bypass (Section~\ref{sec:macroDSE}); Right: SqueezeNet with complex bypass (Section~\ref{sec:macroDSE}).}
	\label{fig:SqueezeNet-architecture}
\end{figure*}


\subsection{The Fire Module}
\label{sec:fire-module}
\vsp

We define the Fire module as follows.
A Fire module is comprised of: a {\em squeeze} convolution layer (which has only 1x1 filters), feeding into an {\em expand} layer that has a mix of 1x1 and 3x3 convolution filters; we illustrate this in Figure~\ref{fig:fire-module}.
The liberal use of 1x1 filters in Fire modules is an application of Strategy 1 from Section~\ref{sec:design-strategies}.
We expose three tunable dimensions (hyperparameters) in a Fire module: $s_{1x1}$, $e_{1x1}$, and $e_{3x3}$.
In a Fire module, $s_{1x1}$ is the number of filters in the squeeze layer (all 1x1), $e_{1x1}$ is the number of 1x1 filters in the expand layer, and $e_{3x3}$ is the number of 3x3 filters in the expand layer.
When we use Fire modules we set $s_{1x1}$ to be less than ($e_{1x1}$ + $e_{3x3}$), so the squeeze layer helps to limit the number of input channels to the 3x3 filters, as  per Strategy 2 from Section~\ref{sec:design-strategies}.

\subsection{The SqueezeNet architecture}
\label{sec:squeezenet-arch}
\vsp

We now describe the SqueezeNet CNN architecture. 
We illustrate in Figure~\ref{fig:SqueezeNet-architecture} that SqueezeNet begins with a standalone convolution layer (conv1), followed by 8 Fire modules (fire2-9), ending with a final conv layer (conv10).
We gradually increase the number of filters per fire module from the beginning to the end of the network.
SqueezeNet performs max-pooling with a stride of 2 after layers conv1, fire4, fire8, and conv10; these relatively late placements of pooling are per Strategy 3 from Section~\ref{sec:design-strategies}.
We present the full SqueezeNet architecture in Table~\ref{T:SqueezeNet-dims}.

\subsubsection{Other SqueezeNet details}
\label{sec:squeezenet-details}
\vspace{-0.1in}

For brevity, we have omitted number of details and design choices about SqueezeNet from Table~\ref{T:SqueezeNet-dims} and Figure~\ref{fig:SqueezeNet-architecture}.
We provide these design choices in the following.
The intuition behind these choices may be found in the papers cited below.

\begin{itemize}
	\setlength\itemsep{0in} 
	\item[$\bullet$]{So that the output activations from 1x1 and 3x3 filters have the same height and width, we add a 1-pixel border of zero-padding in the input data to 3x3 filters of expand modules.}
	\item[$\bullet$]{ReLU~\cite{ReLU} is applied to activations from squeeze and expand layers.}
	\item[$\bullet$]{Dropout~\cite{dropout} with a ratio of 50\% is applied after the fire9 module.}
	\item[$\bullet$]{Note the lack of fully-connected layers in SqueezeNet; this design choice was inspired by the NiN~\cite{NiN} architecture.}
	\item[$\bullet$]{When training SqueezeNet, we begin with a learning rate of 0.04, and we linearly decrease the learning rate throughout training, as described in~\cite{linearLR}.
		For details on the training protocol (e.g. batch size, learning rate, parameter initialization), please refer to our Caffe-compatible configuration files located here: \href{https://github.com/DeepScale/SqueezeNet}{https://github.com/DeepScale/SqueezeNet}.}
	\item[$\bullet$]{The Caffe framework does not natively support a convolution layer that contains multiple filter resolutions (e.g. 1x1 and 3x3)~\cite{jia2014caffe}. To get around this, we implement our expand layer with two separate convolution layers: a layer with 1x1 filters, and a layer with 3x3 filters. Then, we concatenate the outputs of these layers together in the channel dimension. This is numerically equivalent to implementing one layer that contains both 1x1 and 3x3 filters.}
\end{itemize}

We released the SqueezeNet configuration files in the format defined by the Caffe CNN framework. 
However, in addition to Caffe, several other CNN frameworks have emerged, including MXNet~\cite{mxnet}, Chainer~\cite{chainer}, Keras~\cite{keras}, and Torch~\cite{torch}.
Each of these has its own native format for representing a CNN architecture.
That said, most of these libraries use the same underlying computational back-ends such as cuDNN~\cite{cuDNN} and MKL-DNN~\cite{IntelDistributedCNN}.
The research community has ported the SqueezeNet CNN architecture for compatibility with a number of other CNN software frameworks:
\begin{itemize}
	\item MXNet~\cite{mxnet} port of SqueezeNet:~\cite{mxnet-squeezenet}
	
	\item Chainer~\cite{chainer} port of SqueezeNet:~\cite{chainer-squeezenet}
	
	\item Keras~\cite{keras} port of SqueezeNet:~\cite{keras-squeezenet}
	
	\item Torch~\cite{torch} port of SqueezeNet's Fire Modules:~\cite{torch-squeezenet}
	
\end{itemize}

\section{Evaluation of SqueezeNet}
\label{sec:eval-squeezenet}
\vsp

We now turn our attention to evaluating SqueezeNet.
In each of the CNN model compression papers reviewed in Section~\ref{sec:related-model-compression}, the goal was to compress an AlexNet~\cite{alexnet} model that was trained to classify images using the ImageNet~\cite{imagenet} (ILSVRC 2012) dataset.
Therefore, we use AlexNet\footnote{Our baseline is {\tt bvlc\_alexnet} from the Caffe codebase~\cite{jia2014caffe}.} and the associated model compression results as a basis for comparison when evaluating SqueezeNet.

\begin{table}[htb]
	\centering
	\caption[SqueezeNet architectural dimensions]{SqueezeNet architectural dimensions. (The formatting of this table was inspired by the Inception2 paper~\cite{googleBN}.)}
	\label{T:SqueezeNet-dims}
	\includegraphics[width=\textwidth]{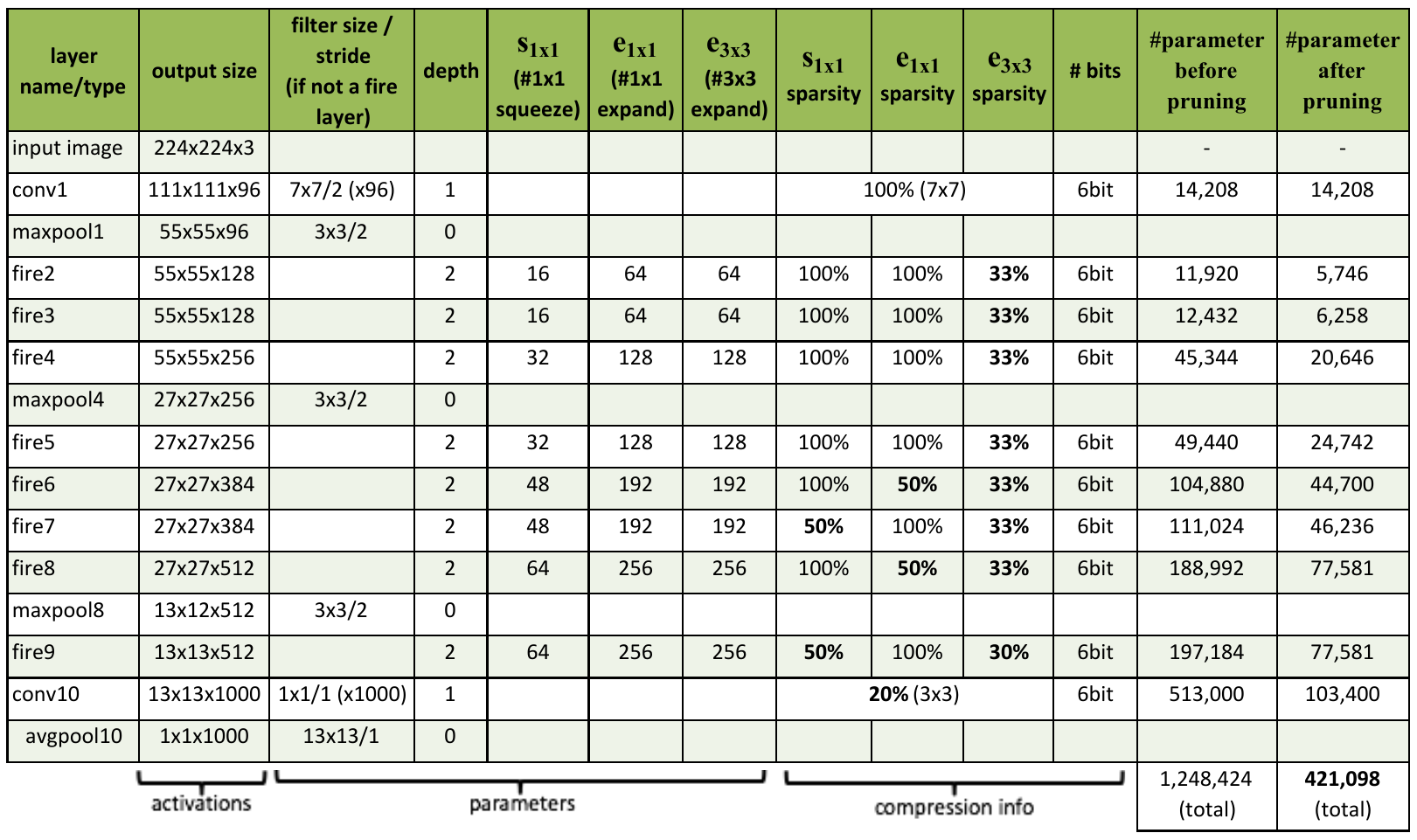}
\end{table}

In Table~\ref{T:model-compression}, we review SqueezeNet in the context of recent model compression results.
The SVD-based approach is able to compress a pretrained AlexNet model by a factor of 5x, while diminishing top-1 accuracy to 56.0\%~\cite{facebook-compress-2014}.
Network Pruning achieves a 9x reduction in model size while maintaining the baseline of 57.2\% top-1 and 80.3\% top-5 accuracy on ImageNet~\cite{dally2015-1}.
Deep Compression achieves a 35x reduction in model size while still maintaining the baseline accuracy level~\cite{dally2015-2}.
Now, with SqueezeNet, we achieve a 50X reduction in model size compared to AlexNet, while {\bf meeting or exceeding the top-1 and top-5 accuracy of AlexNet.}
We summarize all of the aforementioned results in Table~\ref{T:model-compression}.

\begin{table*}[htb]
	\scriptsize
	\caption[Comparing SqueezeNet to model compression approaches]{Comparing SqueezeNet to model compression approaches. By {\em model size}, we mean the number of bytes required to store all of the parameters in the trained model.}
	\label{T:model-compression}
	\centering
	\begin{tabulary}{\textwidth}{|>{\cb}p{2.0cm}|>{\cb}p{2.4cm}|>{\cb}p{0.7cm}|>{\cb}p{2.3cm}|>{\cb}p{1.3cm}|>{\cb}p{1.0cm}|>{\cb}p{1.0cm}|} 
		\hline
		CNN architecture                   & Compression Approach & Data Type & Original $\rightarrow$ Compressed Model Size & Reduction in Model Size vs. AlexNet   & Top-1 ImageNet Accuracy & Top-5 ImageNet Accuracy \\ \hline
		AlexNet & None (baseline) & 32 bit & 240MB   & 1x   & 57.2\% & 80.3\% \\ \hline                                                
		AlexNet & SVD~\cite{facebook-compress-2014} & 32 bit & 240MB $\rightarrow$ 48MB   & 5x   & 56.0\% & 79.4\% \\ \hline 
		AlexNet & Network Pruning~\cite{dally2015-1} & 32 bit & 240MB $\rightarrow$  27MB   & 9x   & 57.2\% & 80.3\% \\ \hline 
		AlexNet & Deep Compression~\cite{dally2015-2} & 5-8 bit & 240MB $\rightarrow$  6.9MB   & 35x   & 57.2\% & 80.3\% \\ \hline 
		SqueezeNet (ours) & None & 32 bit & 4.8MB & {\bf 50x}   & 57.5\% & 80.3\% \\ \hline 
		SqueezeNet (ours) & Deep Compression & 8 bit & 4.8MB $\rightarrow$ 0.66MB & {\bf 363x}  & 57.5\% & 80.3\% \\ \hline 
		SqueezeNet (ours) & Deep Compression & 6 bit & 4.8MB $\rightarrow$ 0.47MB & {\bf 510x}  & 57.5\% & 80.3\% \\ \hline 
		
	\end{tabulary}
\end{table*}

It appears that we have surpassed the state-of-the-art results from the model compression community:
even when using uncompressed 32-bit values to represent the model, SqueezeNet has a $1.4\times$ smaller model size than the best efforts from the model compression community while maintaining or exceeding the baseline accuracy.
Until now, an open question has been: {\em are small models amenable to compression, or do small models ``need" all of the representational power afforded by dense floating-point values?}
To find out, we applied Deep Compression~\cite{dally2015-2} to SqueezeNet, using 33\% sparsity\footnote{Note that, due to the storage overhead of storing sparse matrix indices, 33\% sparsity leads to somewhat less than a $3\times$ decrease in model size.} and 8-bit quantization.
This yields a 0.66 MB model ($363\times$ smaller than 32-bit AlexNet) with equivalent accuracy to AlexNet.
Further, applying Deep Compression with 6-bit quantization and 33\% sparsity on SqueezeNet, we produce a 0.47MB model ($510\times$ smaller than 32-bit AlexNet) with equivalent accuracy.
{\bf Our small model is indeed amenable to compression.}

In addition, these results demonstrate that Deep Compression~\cite{dally2015-2} not only works well on CNN architectures with many parameters (e.g. AlexNet and VGG), but it is also able to compress the already compact, fully convolutional SqueezeNet architecture. 
Deep Compression compressed SqueezeNet by $10\times$ while preserving the baseline accuracy.
In summary: by combining CNN architectural innovation (SqueezeNet) with state-of-the-art compression techniques (Deep Compression), we achieved a $510\times$ reduction in model size with no decrease in accuracy compared to the baseline.

Finally, note that Deep Compression~\cite{dally2015-1} uses a {\em codebook} as part of its scheme for quantizing CNN parameters to 6- or 8-bits of precision.
Therefore, on most commodity processors, it is {\em not} trivial to achieve a speedup of $\frac{32}{8}=4x$ with 8-bit quantization or $\frac{32}{6}=5.3x$ with 6-bit quantization using the scheme developed in Deep Compression.
However, Han \etal developed custom hardware -- {\em Efficient Inference Engine (EIE)} -- that can compute codebook-quantized CNNs more efficiently~\cite{EIE}.
In addition, in the months since we released SqueezeNet, P. Gysel developed a strategy called {\em Ristretto} for linearly quantizing SqueezeNet to 8 bits~\cite{Ristretto}.
Specifically, Ristretto does computation in 8 bits, and it stores parameters and activations in 8-bit data types.
Using the Ristretto strategy for 8-bit computation in SqueezeNet inference, Gysel observed less than 1 percentage-point of drop in accuracy when using 8-bit instead of 32-bit data types.

\section{CNN Microarchitecture Design Space Exploration}
\label{sec:microDSE}
\vsp

So far, we have proposed architectural design strategies for small models, followed these principles to create SqueezeNet, and discovered that SqueezeNet is 50x smaller than AlexNet with equivalent accuracy.
However, SqueezeNet and other models reside in a broad and largely unexplored design space of CNN architectures.
Now, in Sections~\ref{sec:microDSE} and~\ref{sec:macroDSE}, we explore several aspects of the design space. 
We divide this architectural exploration into two main topics: {\em microarchitectural exploration} (per-module layer dimensions and configurations) and {\em macroarchitectural exploration} (high-level end-to-end organization of modules and other layers). 

In this section, we design and execute experiments with the goal of providing intuition about the shape of the microarchitectural design space with respect to the design strategies that we proposed in Section~\ref{sec:design-strategies}.
Note that our goal here is {\em not} to maximize accuracy in every experiment, but rather to understand the impact of CNN architectural choices on model size and accuracy.

\subsection{CNN Microarchitecture metaparameters}
\label{sec:metaparameters}
\vsp

In SqueezeNet, each Fire module has three dimensional hyperparameters that we defined in Section~\ref{sec:fire-module}: $s_{1x1}$, $e_{1x1}$, and $e_{3x3}$. 
SqueezeNet has 8 Fire modules with a total of 24 dimensional hyperparameters.
To do broad sweeps of the design space of SqueezeNet-like architectures, we define the following set of higher level {\em metaparameters} which control the dimensions of all Fire modules in a CNN.
We define $base_e$ as the number of {\em expand} filters in the first Fire module in a CNN.
After every $freq$ Fire modules, we increase the number of expand filters by $incr_e$.
In other words, for Fire module $i$, the number of expand filters is $e_i=base_e + (incr_e*{\left\lfloor{\frac{i}{freq}}\right\rfloor}$).
In the expand layer of a Fire module, some filters are 1x1 and some are 3x3; we define $e_i = e_{i,{1x1}} + e_{i,{3x3}}$ with $pct_{3x3}$ (in the range $[0,1]$, shared over all Fire modules) as the percentage of expand filters that are 3x3.
In other words, $e_{i,{3x3}} = e_i*pct_{3x3}$, and $e_{i,{1x1}} = e_i*(1-pct_{3x3})$. 
Finally, we define the number of filters in the squeeze layer of a Fire module using a metaparameter called the {\em squeeze ratio (SR)} (again, in the range $[0,1]$, shared by all Fire modules): $s_{i,{1x1}} = SR * e_i$ (or equivalently $s_{i,{1x1}} = SR * (e_{i,{1x1}} + e_{i,{3x3}})$).
SqueezeNet (Table~\ref{T:SqueezeNet-dims}) is an example architecture that we generated with the aforementioned set of metaparameters.
Specifically, SqueezeNet has the following metaparameters: $base_e = 128$, $incr_e = 128$, $pct_{3x3} = 0.5$, $freq=2$, and $SR = 0.125$.

\subsection{Squeeze Ratio}
\label{sec:SR}
\vsp

In Section~\ref{sec:design-strategies}, we proposed decreasing the number of parameters by using {\em squeeze layers} to decrease the number of input channels seen by 3x3 filters.
We defined the {\em squeeze ratio (SR)} as the ratio between the number of filters in {\em squeeze} layers and the number of filters in {\em expand} layers.
We now design an experiment to investigate the effect of the squeeze ratio on model size and accuracy.

\begin{figure}[!t]
	\centering
	\subfigure[ Exploring the impact of the squeeze ratio~($SR$) on model size and accuracy.]
	{
		\includegraphics[height=0.35\textwidth]{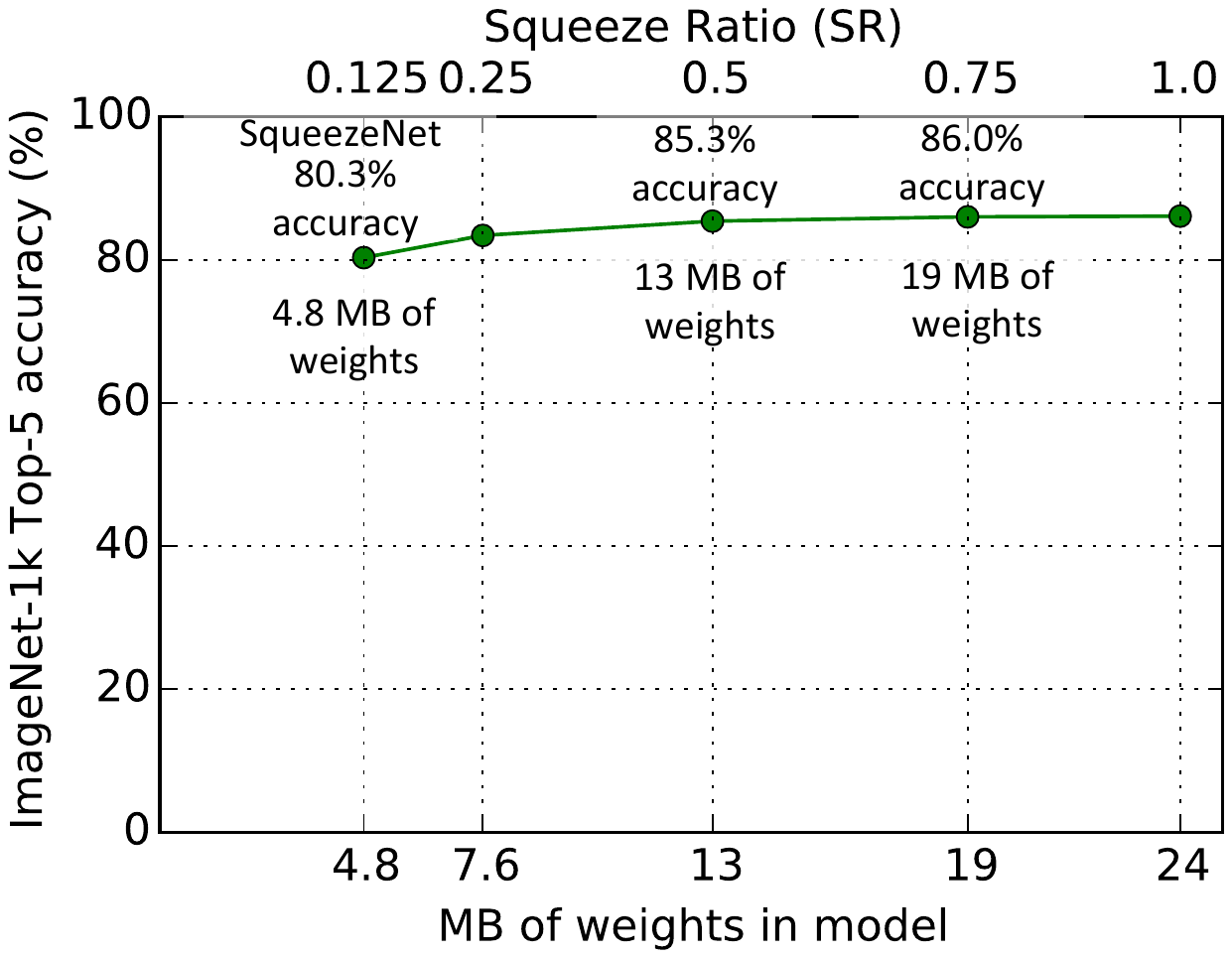}
		\label{fig:squeeze-ratio}
	}
	\hspace{5pt}
	\subfigure[Exploring the impact of the ratio of 3x3 filters in expand layers~($pct_{3x3}$) on model size and accuracy.]
	{
		\includegraphics[height=0.35\textwidth]{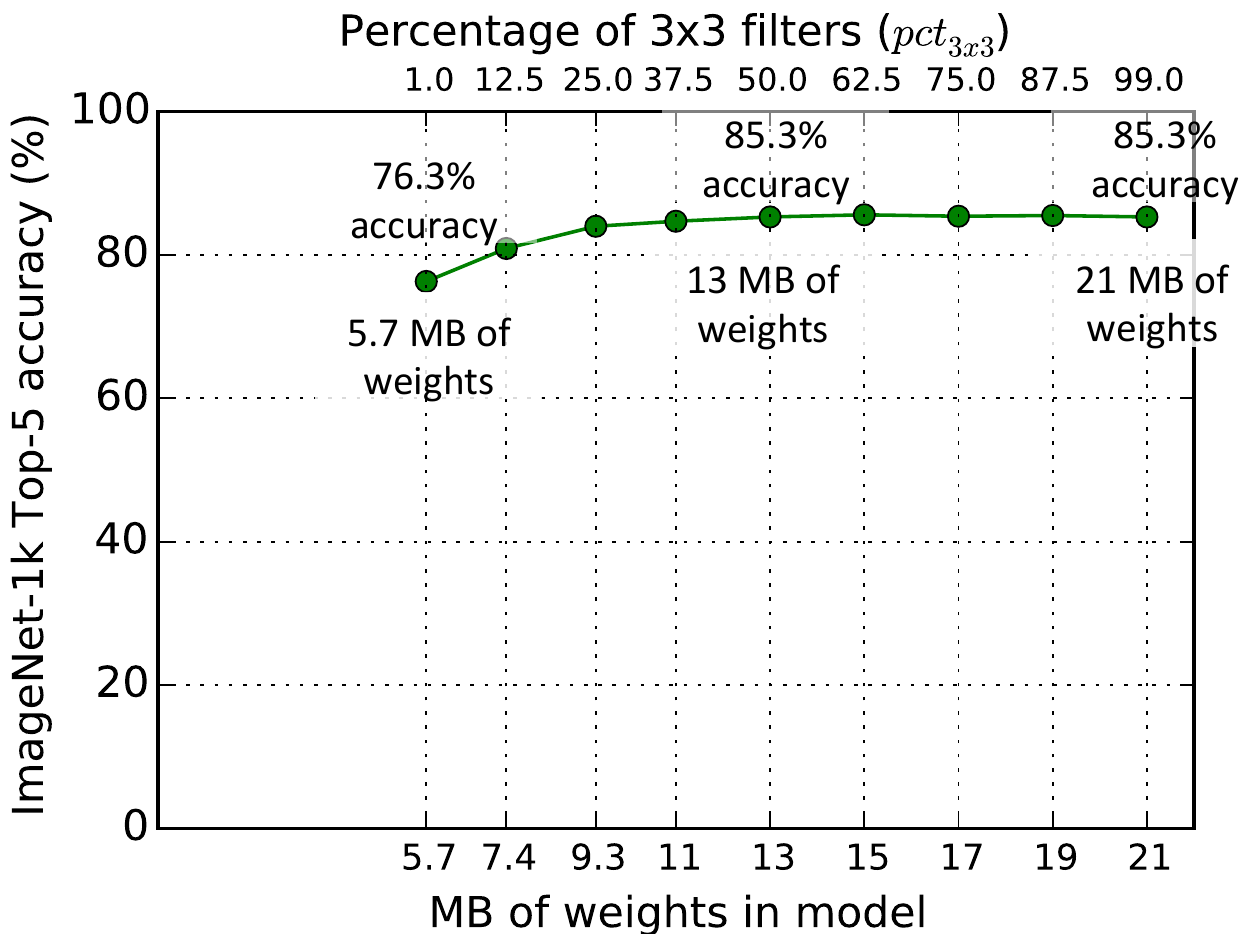}
		\label{fig:pct_3x3}
	}
	\label{fig:macroDSE}
	\caption{Microarchitectural design space exploration.}
\end{figure}

In these experiments, we use SqueezeNet (Figure~\ref{fig:SqueezeNet-architecture}) as a starting point.
As in SqueezeNet, these experiments use the following metaparameters: $base_e = 128$, $incr_e = 128$, $pct_{3x3} = 0.5$, and $freq=2$.
We train multiple models, where each model has a different squeeze ratio (SR)\footnote{Note that, for a given model, all Fire layers share the same squeeze ratio.} in the range [0.125, 1.0].
In Figure~\ref{fig:squeeze-ratio}, we show the results of this experiment, where each point on the graph is an independent model that was trained from scratch.
SqueezeNet is the SR=0.125 point in this figure.\footnote{Note that we named it {\em Squeeze}Net because it has a low squeeze ratio (SR). That is, the squeeze layers in SqueezeNet have 0.125x the number of filters as the expand layers.}
From this figure, we learn that increasing SR beyond 0.125 can further increase ImageNet top-5 accuracy from 80.3\% (i.e. AlexNet-level) with a 4.8MB model to 86.0\% with a 19MB model.
Accuracy plateaus at 86.0\% with SR=0.75 (a 19MB model), and setting SR=1.0 further increases model size without improving accuracy.

\subsection{Trading off 1x1 and 3x3 filters}
\label{sec:pct_3x3}
\vsp

In Section~\ref{sec:design-strategies}, we proposed decreasing the number of parameters in a CNN by replacing some 3x3 filters with 1x1 filters.
An open question is, {\em how important is spatial resolution in CNN filters?}
The VGG~\cite{VGG-19} architectures have 3x3 spatial resolution in most layers' filters; GoogLeNet~\cite{googlenet} and Network-in-Network (NiN)~\cite{NiN} have 1x1 filters in some layers.
In GoogLeNet and NiN, the authors simply propose a specific quantity of 1x1 and 3x3 filters without further analysis.\footnote{To be clear, each filter is 1x1xChannels or 3x3xChannels, which we abbreviate to 1x1 and 3x3.}
Here, we attempt to shed light on how the proportion of 1x1 and 3x3 filters affects model size and accuracy.

We use the following metaparameters in this experiment: $base_e = incr_e = 128$, $freq=2$, $SR=0.500$, and we vary $pct_{3x3}$ from 1\% to 99\%. 
In other words, each Fire module's expand layer has a predefined number of filters partitioned between 1x1 and 3x3, and here we turn the knob on these filters from ``mostly 1x1'' to ``mostly 3x3''.
As in the previous experiment, these models have 8 Fire modules, following the same organization of layers as in Figure~\ref{fig:SqueezeNet-architecture}.
We show the results of this experiment in Figure~\ref{fig:pct_3x3}.
Note that the 13MB models in Figure~\ref{fig:squeeze-ratio} and Figure~\ref{fig:pct_3x3} are the same architecture: $SR=0.500$ and $pct_{3x3}=50\%$.
We see in Figure~\ref{fig:pct_3x3} that the top-5 accuracy plateaus at 85.6\% using 50\% 3x3 filters, and further increasing the percentage of 3x3 filters leads to a larger model size but provides no improvement in accuracy on ImageNet.

\section{CNN Macroarchitecture Design Space Exploration}
\label{sec:macroDSE}
\vsp

So far we have explored the design space at the microarchitecture level, i.e. the contents of individual modules of the CNN.
Now, we explore design decisions at the macroarchitecture level concerning the high-level connections among Fire modules.
Inspired by ResNet~\cite{resnet}, we explored three different architectures: 

\vbox{
\begin{itemize}
	\item[$\bullet$]{Vanilla SqueezeNet (as per the prior sections).}
	\item[$\bullet$]{SqueezeNet with simple bypass connections between some Fire modules. (Inspired by~\cite{highway-networks,resnet}.)}
	\item[$\bullet$]{SqueezeNet with complex bypass connections between the remaining Fire modules.}
\end{itemize}
}

We illustrate these three variants of SqueezeNet in Figure~\ref{fig:SqueezeNet-architecture}.

Our {\em simple bypass} architecture adds bypass connections around Fire modules 3, 5, 7, and 9, requiring these modules to learn a residual function between input and output.
As in ResNet, to implement a bypass connection around Fire3, we set the input to Fire4 equal to (output of Fire2 + output of Fire3), where the + operator is elementwise addition.
This changes the regularization applied to the parameters of these Fire modules, and, as per ResNet, can improve the final accuracy and/or ability to train the full model.

One limitation is that, in the straightforward case, the number of input channels and number of output channels has to be the same; as a result, only half of the Fire modules can have simple bypass connections, as shown in the middle diagram of Fig~\ref{fig:SqueezeNet-architecture}.
When the ``same number of channels" requirement can't be met, we use a {\em complex bypass} connection, as illustrated on the right of Figure~\ref{fig:SqueezeNet-architecture}. 
While a simple bypass is ``just a wire," we define a complex bypass as a bypass that includes a 1x1 convolution layer with the number of filters set equal to the number of output channels that are needed.
Note that complex bypass connections add extra parameters to the model, while simple bypass connections do not.

In addition to changing the regularization, it is intuitive to us that adding bypass connections would help to alleviate the representational bottleneck introduced by squeeze layers.
In SqueezeNet, the squeeze ratio (SR) is 0.125, meaning that every squeeze layer has 8x fewer output channels than the accompanying expand layer.
Due to this severe dimensionality reduction, a limited amount of information can pass through squeeze layers.
However, by adding bypass connections to SqueezeNet, we open up avenues for information to flow {\em around} the squeeze layers.

\begin{table}[t]
	\centering
	\caption{SqueezeNet accuracy and model size using different macroarchitecture configurations}
	\label{table:macroarchitecture}
	\footnotesize
	\begin{tabulary}{0.9\textwidth}{|C|C|C|C|}
		\hline
		Architecture                & Top-1 Accuracy & Top-5 Accuracy & Model Size \\ \hline
		Vanilla SqueezeNet          & 57.5\%         & 80.3\%         & 4.8MB      \\ \hline
		SqueezeNet + Simple Bypass  & \bf{60.4}\%    & \bf{82.5}\%         & 4.8MB      \\ \hline
		SqueezeNet + Complex Bypass & 58.8\%         & 82.0\%           & 7.7MB     \\ \hline
	\end{tabulary}
\end{table}

We trained SqueezeNet with the three macroarchitectures in Figure~\ref{fig:SqueezeNet-architecture} and compared the accuracy and model size in Table~\ref{table:macroarchitecture}. 
We fixed the microarchitecture to match SqueezeNet as described in Table~\ref{T:SqueezeNet-dims} throughout the macroarchitecture exploration. 
Complex and simple bypass connections both yielded an accuracy improvement over the vanilla SqueezeNet architecture. Interestingly, the simple bypass enabled a higher accuracy accuracy improvement than complex bypass.
Adding the simple bypass connections yielded an increase of 2.9 percentage-points in top-1 accuracy and 2.2 percentage-points in top-5 accuracy without increasing model size.

\section{Conclusions}
\label{sec:conclusions}
\vsp

In this paper, we have proposed steps toward a more disciplined approach to the design-space exploration of convolutional neural networks. 
Toward this goal we have presented SqueezeNet, a CNN architecture that has $50\times$ fewer parameters than AlexNet and maintains AlexNet-level accuracy on ImageNet.
We also compressed SqueezeNet to less than 0.5MB, or $510\times$ smaller than AlexNet without compression. 
Since we released this paper as a technical report in 2016, Song Han and his collaborators have experimented further with SqueezeNet and model compression.
Using a new approach called {\em Dense-Sparse-Dense (DSD)}~\cite{DSD}, Han \etal use model compression during training as a regularizer to further improve accuracy, producing a compressed set of SqueezeNet parameters that is 1.2 percentage-points more accurate on ImageNet-1k, and also producing an uncompressed set of SqueezeNet parameters that is 4.3 percentage-points more accurate, compared to our results in Table~\ref{T:model-compression}.

We mentioned near the beginning of this paper that small models are more amenable to on-chip implementations on FPGAs.
Since we released the SqueezeNet model, Gschwend has developed a variant of SqueezeNet and implemented it on an FPGA~\cite{ZynqNet}.
As we anticipated, Gschwend was able to able to store the parameters of a SqueezeNet-like model entirely within the FPGA and eliminate the need for off-chip memory accesses to load model parameters.

In the context of this paper, we focused on ImageNet as a target dataset.
However, it has become common practice to apply ImageNet-trained CNN representations to a variety of applications such as fine-grained object recognition~\cite{DPD, decaf}, logo identification in images~\cite{DeepLogo}, and generating sentences about images~\cite{forrestMicrosoft}.
ImageNet-trained CNNs have also been applied to a number of applications pertaining to autonomous driving, including pedestrian and vehicle detection in images~\cite{DenseNet,DPMareCNN,Ashraf2016} and videos~\cite{Urtasun2015}, as well as segmenting the shape of the road~\cite{SegNet}. 
We think SqueezeNet will be a good candidate CNN architecture for a variety of applications, especially those in which small model size is of importance.

SqueezeNet is one of several new CNNs that we have discovered while broadly exploring the design space of CNN architectures.
We hope that SqueezeNet will inspire the reader to consider and explore the broad range of possibilities in the design space of CNN architectures and to perform that exploration in a more systematic manner.



\bibliography{bibliography}
\bibliographystyle{iclr2017_conference}

\end{document}